\documentclass[lettersize,journal]{IEEEtran}
\usepackage{amsmath,amsfonts}
\usepackage{algorithmic}
\usepackage{algorithm}
\usepackage{array}
\usepackage[caption=false,font=normalsize,labelfont=sf,textfont=sf]{subfig}
\usepackage{textcomp}
\usepackage{stfloats}
\usepackage{url}
\usepackage{verbatim}
\usepackage{graphicx}
\usepackage{cite}

\usepackage[table,xcdraw]{xcolor}
\usepackage{cleveref}
\usepackage{multirow}
\usepackage{amssymb}
\usepackage{wrapfig}
\usepackage{booktabs}

\begin{document}

\title{Biologically Inspired Spiking Diffusion Model with \\ Adaptive Lateral Selection Mechanism}

\author{Linghao Feng$^{1,2,*}$, Dongcheng Zhao$^{1,3,*}$, Sicheng Shen$^{1,2}$, Yi Zeng$^{1,2,3,4,\dagger}$
\thanks{$^1$Brain-inspired Cognitive Intelligence Lab, Institute of Automation, Chinese Academy of Sciences.}
\thanks{$^2$School of Future Technology, University of Chinese Academy of Sciences.}
\thanks{$^3$Center for Long-term Artificial Intelligence}
\thanks{$^4$State Key Laboratory of Brain Cognition and Brain-inspired Intelligence Technology}
\thanks{$*$ Equal Contribution. $\dagger$ Corresponding Author}
}

\markboth{Journal of \LaTeX\ Class Files,~Vol.~14, No.~8, August~2021}%
{Shell \MakeLowercase{\textit{et al.}}: A Sample Article Using IEEEtran.cls for IEEE Journals}


\maketitle

\begin{abstract}
Lateral connection is a fundamental feature of biological neural circuits, facilitating local information processing and adaptive learning. In this work, we integrate lateral connections with a substructure selection network to develop a novel diffusion model based on spiking neural networks (SNNs). Unlike conventional artificial neural networks, SNNs employ an intrinsic spiking inner loop to process sequential binary spikes. We leverage this spiking inner loop alongside a lateral connection mechanism to iteratively refine the substructure selection network, enhancing model adaptability and expressivity. Specifically, we design a lateral connection framework comprising a learnable lateral matrix and a lateral mapping function, both implemented using spiking neurons, to dynamically update lateral connections. Through mathematical modeling, we establish that the proposed lateral update mechanism, under a well-defined local objective, aligns with biologically plausible synaptic plasticity principles. Extensive experiments validate the effectiveness of our approach, analyzing the role of substructure selection and lateral connection during training. Furthermore, quantitative comparisons demonstrate that our model consistently surpasses state-of-the-art SNN-based generative models across multiple benchmark datasets.
\end{abstract}


\section{Introduction}
\IEEEPARstart{S}{piking} Spiking Neural Networks (SNNs)~\cite{maass1997networks} constitute an advanced computational paradigm that more accurately emulates the functional principles of neural processing in the human brain compared to conventional artificial neural networks (ANNs). Unlike ANNs, which rely on continuous-valued activations, SNNs encode and transmit information through discrete binary spikes, closely aligning with the spike-based communication observed in biological neural circuits. This event-driven mechanism not only enhances the biological plausibility of SNNs but also introduces a sparse temporal coding scheme, which is particularly advantageous for energy-efficient computation. When deployed on neuromorphic hardware platforms~\cite{roy2019towards}, SNNs can leverage this sparsity to achieve significantly reduced energy consumption, mirroring the remarkable energy efficiency of the human brain during cognitive inference tasks.

A fundamental challenge in applying SNNs to machine learning tasks lies in the development of effective optimization and learning methodologies for spiking neurons. Classical spiking neuron models, such as the Leaky Integrate-and-Fire (LIF) model~\cite{liu2001spike}, generate binary spike outputs (0 or 1) based on whether their membrane potential surpasses a predefined threshold. However, this inherently non-differentiable spiking mechanism poses significant challenges for conventional gradient-based optimization algorithms. Traditional biologically inspired approaches, such as Spike-Timing-Dependent Plasticity (STDP)~\cite{bi1998synaptic,dong2023unsupervised}, leverage the relative firing times of neurons to guide synaptic weight updates. While STDP offers strong biological plausibility, its application is often constrained to shallow networks and simpler tasks, limiting its scalability to deeper architectures and more complex machine learning scenarios. To address these limitations and enhance the performance of SNN-based models on more sophisticated downstream tasks, the surrogate gradient method~\cite{wu2018spatio,zhang2021rectified} has emerged as a promising solution. This technique maintains the step-function behavior of spiking neurons during the forward pass while introducing a differentiable surrogate function during backpropagation to approximate gradient computation. The integration of surrogate gradients enables SNNs to be trained with standard gradient-based algorithms, significantly broadening their applicability to advanced tasks, including object detection~\cite{kim2020spiking}, object tracking~\cite{zhang2022spiking}, voice activity detection~\cite{martinelli2020spiking}, and language models~\cite{zhu2023spikegpt,shen2023astrocyte}. This methodological advancement has paved the way for SNNs to demonstrate competitive performance in diverse and challenging machine learning applications.
\begin{figure}[t]
    \centering
    \includegraphics[width=1.0\columnwidth]{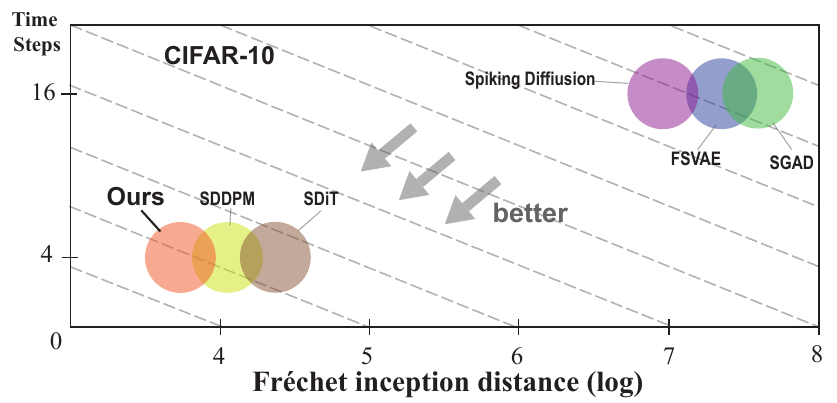}
    \caption{Comparison of the proposed method and other SNN-based generative models on CIFAR-10 in terms of FID and spiking time steps}
    \label{fig:compare_fids}
\end{figure}
Generative models have emerged as a pivotal focus in artificial intelligence (AI) and deep learning, capturing widespread attention within the research community. These models are designed to learn complex probability distributions through neural networks and generate novel samples by drawing from these distributions. In natural language processing, transformer-based and autoregressive models, such as GPT~\cite{achiam2023gpt}, have demonstrated remarkable human-like linguistic proficiency. In the realm of computer vision, state-of-the-art generative models like Stable Diffusion~\cite{rombach2021highresolution} and DALL·E~\cite{ramesh2022hierarchical} have achieved unprecedented success in generating high-quality images from textual prompts, closely emulating human artistic creativity. Recent studies in neuroscience suggest that the brain engages in generative processes during activities such as creative thinking, visualization, and even dreaming~\cite{Bucci2017-BUCSAD-3}. These processes involve generating novel ideas or reconstructing sensory experiences through predictive coding and generative recall mechanisms~\cite{huang2011predictive}. This alignment underscores the natural synergy between generative models in machine learning and brain-inspired computational principles, providing a compelling rationale for their integration. SNNs which mimic the brain’s spike-based information transmission, provide a biologically plausible foundation for constructing generative models when combined with other neuromorphic principles. Inspired by this perspective, recent studies have integrated SNNs with various generative models, including variational autoencoders (VAEs), generative adversarial networks (GANs), and diffusion models~\cite{kamata2022fully,kotariya2022spiking,feng2024spiking,cao2024spiking,liu2023spiking}. These approaches demonstrate the versatility of SNNs in generative tasks and highlight their potential for robust performance across complex domains.

In this work, we develop a spiking diffusion model that integrates a biologically inspired sub-structure selection mechanism within a diffusion framework, leveraging a spiking-form transformer~\cite{zhou2022spikformer}. Our approach draws inspiration from lateral connections in biological neural circuits~\cite{pinotsis2011neural, angelucci2006contribution}, analogous to the Mixture of Experts (MoE) paradigm~\cite{shazeer2017outrageously, fedus2022switch, zuo2021taming, park2024switch}. However, unlike traditional MoE, where expert selection is explicitly predefined, our sub-structure selection module dynamically facilitates information flow through lateral connections, enhancing adaptability and computational efficiency. Our SNN-based denoising model distinguishes between the "outer loop" of the reverse diffusion process, which iteratively refines generated samples, and the "inner loop" of spiking neuron dynamics, where sequential spiking activity iteratively adjusts lateral connections to improve generative performance. This design enables adaptive learning within the diffusion framework, facilitating structured information processing akin to biological neural systems. 

To support this design, we establish a rigorous mathematical framework that characterizes neuronal activity through latent variables and formulates inter-population interactions as constrained optimization problems. Under the assumption of a reasonable local loss function, we demonstrate that the update process of lateral connections approximates the synaptic plasticity mechanism governed by STDP, reinforcing the biological plausibility of our approach. 
Specifically, we introduce a lateral aggregation matrix to modulate the propagation of neuronal activity across lateral pathways and a spike mapping function that ensures information transmission remains consistent with spiking constraints. Theoretical analysis provides guarantees that our approach effectively adjusts lateral information flow based on spiking patterns, thereby capturing long-range temporal dependencies and optimizing generative performance in complex tasks. We validate the effectiveness of our method through extensive experiments on standard benchmark datasets, including MNIST, CelebA~\cite{liu2015faceattributes}, and CIFAR-10. Experimental results demonstrate that our model consistently surpasses existing SNN-based generative approaches, achieving superior Fréchet Inception Distance (FID) scores. As illustrated in Figure~\ref{fig:compare_fids}, our method outperforms competing approaches in both generative quality and spiking efficiency, highlighting its advantages in structured probabilistic modeling. In summary, the contributions of our work are as follows:
\begin{enumerate}
    \item  We introduce a novel spiking diffusion model that incorporates a biologically inspired sub-structure selection network alongside a lateral connection mechanism, effectively capturing the dynamic information processing observed in biological neural systems.
    \item Through rigorous mathematical modeling, we establish that, under a well-defined local objective, updating lateral parameters via surrogate gradients closely approximates biologically plausible synaptic plasticity mechanisms. This theoretical foundation informs the design of an adaptive and computationally efficient lateral connection update rule, ensuring stable and effective learning.  
    \item  We conduct extensive empirical evaluations on multiple benchmark datasets, consistently demonstrating superior performance over existing SNN-based generative models. Furthermore, ablation studies confirm the crucial role of sub-structure selection and lateral connectivity mechanisms, highlighting their contributions to enhancing model efficiency and robustness during training.  
\end{enumerate}

\section{Related Work}
\subsection{SNN-based Generative Models}
Recent advances in SNNs have led to the development of a variety of generative models, demonstrating their potential in diverse machine learning tasks. The Fully Spiking Variational Autoencoder (FSVAE)~\cite{kamata2022fully} employs fully spiking neurons and adapts the latent variable distribution from a Gaussian to a Bernoulli distribution, optimizing the model for spike-based signal processing. And~\cite{feng2024time} constructed a codebook that mimics the time cells in the hippocampus, capable of recording temporal information, and designed a VQ-VAE based on SNN. 
Spiking-GAN~\cite{kotariya2022spiking} pioneered the application of SNNs within a GAN~\cite{goodfellow2020generative} framework, utilizing the TTFS~\cite{oh2020hardware} encoding method to facilitate spike-driven learning. To address domain and time-step inconsistencies in SNN-based generative models, SGAD~\cite{feng2024spiking} introduces novel strategies, significantly enhancing GAN performance. In the realm of diffusion models, Spiking-Diffusion~\cite{liu2023spiking} presents an SNN-based vector-quantized variational autoencoder (VQ-VAE)~\cite{van2017neural} and integrates a fully spiking discrete denoising diffusion probabilistic model (DDPM)~\cite{austin2021structured} to improve codebook sampling. Similarly, SDDPM~\cite{cao2024spiking} adapts the classical DDPM framework~\cite{ho2020denoising} to an SNN architecture, achieving competitive generative performance. 
In contrast to these approaches, our method introduces a biologically inspired sub-structure selection mechanism combined with a lateral connection and propagation strategy. 
This integration not only aligns more closely with the biological plausibility of SNNs but also enhances dynamic information flow within the network. Empirical evaluations demonstrate that our model consistently outperforms existing SNN-based generative models, validating the effectiveness of integrating brain-inspired mechanisms into the generative modeling framework.
\subsection{Diffusion Models}

Diffusion models are generative frameworks characterized by a forward diffusion process, which progressively adds noise to data, and a reverse denoising process aimed at reconstructing the original stimuli. The Denoising Diffusion Probabilistic Model (DDPM)~\cite{ho2020denoising} pioneered this approach using a U-Net architecture to estimate noise and recover original inputs. Building on this foundation, Score-Based Generative Models~\cite{song2020score} provided a score-matching perspective, extending the diffusion process to the continuous-time domain. Further advancements in diffusion modeling include the Denoising Diffusion Implicit Model (DDIM)~\cite{song2020denoising}, which relaxed the Markov assumption of the forward diffusion process, proposing a non-Markovian framework that significantly accelerates sampling. The Diffusion Transformer (DiT)~\cite{peebles2023scalable} introduced a transformer-based architecture to replace the traditional U-Net, demonstrating enhanced generative performance. This architecture was further improved by integrating MoE strategies~\cite{park2024switch}, optimizing model efficiency and accuracy. In conditional generation, Dhariwal and Nichol~\cite{dhariwal2021diffusion} utilized a classifier-based guidance technique, which was later refined by Ho et al.~\cite{ho2022classifier} to eliminate the need for an external classifier by leveraging the model's intrinsic null condition. StableDiffusion\cite{rombach2022high} extended the diffusion framework to the latent space, integrating natural language processing capabilities to facilitate high-quality text-to-image generation.

\begin{figure*}[t]
    \centering
    \includegraphics[width=0.9\textwidth]{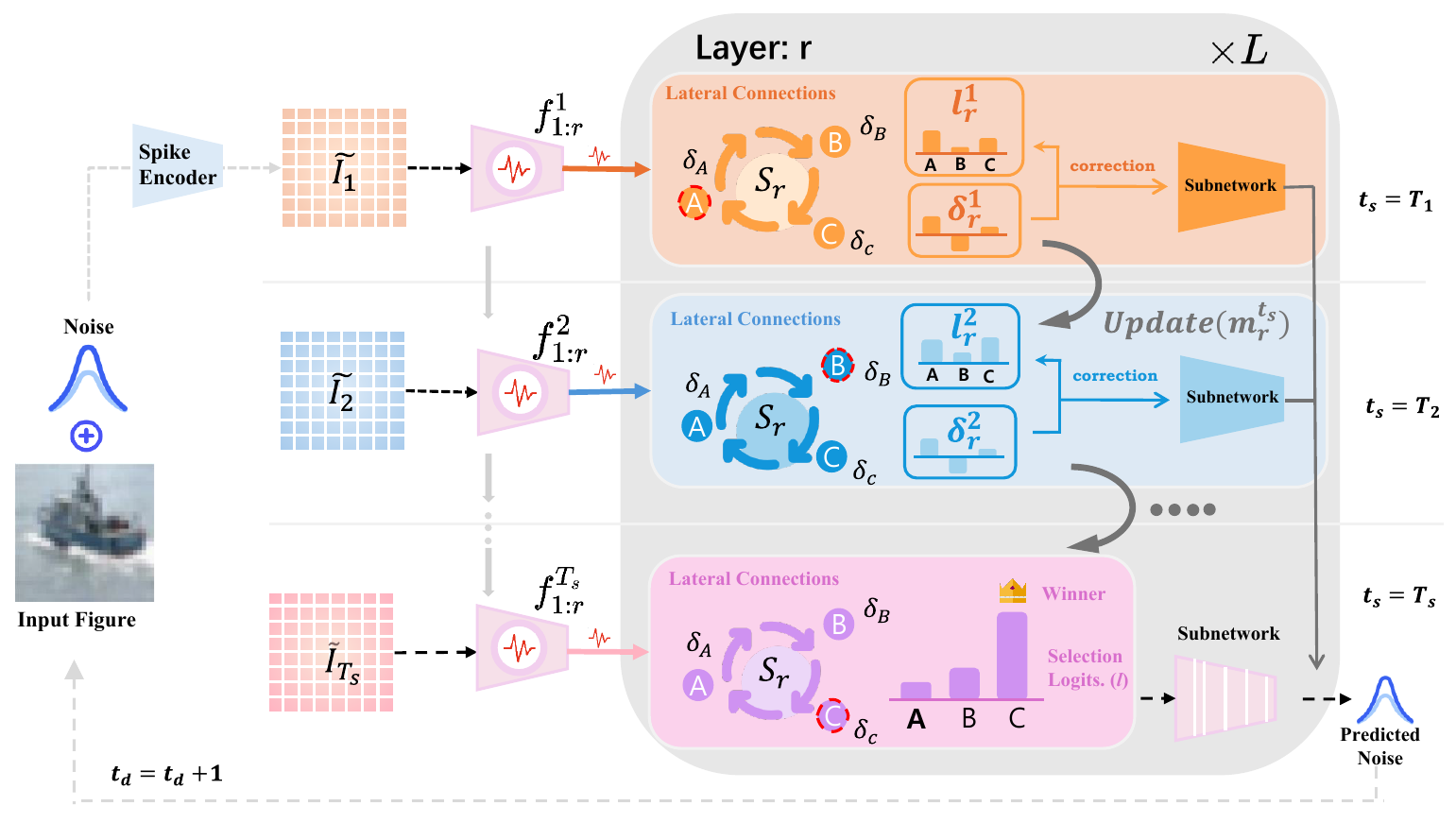}
    \caption{Overview of the proposed lateral connection spiking diffusion. \( t_s = 1, 2 \dots, T_s \) denote the inner cycles of the spiking neurons. \( f^{t_s}_{1:r} \) represents the portion of the entire model before the \( r \)-th substructure selection module at time step $t_s$, while \( l^{t_s}_r \) and \( \delta^{t_s}_r \) denote the initial logits to the \( r \)-th substructure selection module and the correction value at time step \( t_s \), respectively. The lateral connections output \( \delta^{t_s}_r \) and update the membrane potential \( m^{t_s}_r \); then, \( \delta^{t_s}_r \) is used to correct the logits \( l^{t_s}_r \), resulting in the selected substructure network.}
    \label{fig:pipeline}
\end{figure*}

\section{Method}

\subsection{Overview}
In this section, we present a spiking diffusion framework that integrates a transformer-based denoising network~\cite{ho2020denoising} with LIF neurons. The model incorporates a lateral aggregation matrix $\Lambda_{agg}$ and a spike mapping function $\Omega$ to dynamically refine the substructure selection during the denoising process. Our architecture employs an outer diffusion loop for denoising and an inner spiking loop to update the selection module through lateral connection. Through mathematical modeling, we show that under specific local objectives, this approach approximates the biologically plausible learning rules of STDP~\cite{bengio2015stdp}. This design enhances generative performance while maintaining biological plausibility. The schematic overview of the pipeline is shown in Figure~\ref{fig:pipeline}.

\subsection{Leaky Integrate-and-Fire Model} \label{sec:lif}
The Leaky Integrate-and-Fire (LIF) neuron model~\cite{liu2001spike} is employed as the fundamental computational unit in our SNN due to its computational efficiency and its biologically plausible abstraction of neuronal spiking dynamics. The discrete-time dynamics of the LIF neuron, encompassing membrane potential accumulation and spiking behavior, are defined by:
\begin{align}
	&H^t = V^{t-1} + \frac{1}{\tau} \cdot \left( I^t - \left( V^{t-1} - V_{\text{reset}} \right) \right) \\
	&S^t = \Theta \left( H^t - V_{\text{th}} \right) \\
	&V^t = H^t \left( 1 - S^t \right) + S^t \cdot V_{\text{reset}}
	\end{align}
where \( \tau \) is the membrane time constant, \( I^t \) represents the input synaptic current, and \( H^t \) denotes the membrane potential prior to spike generation. A spike \( S^t \) is emitted when the membrane potential exceeds the threshold \( V_{\text{th}} \),triggering a reset to \( V_{\text{reset}} \). The Heaviside step function \( \Theta(v) \) is defined as 1 when \( v \geq 0 \), and 0 otherwise. Given the non-differentiable nature of the Heaviside function, we apply the surrogate gradient method~\cite{neftci2019surrogate} to enable gradient-based optimization within the SNN framework.

\subsection{Spiking Diffusion with Lateral Connection} \label{sec:model}

In this section, we present the implementation of the proposed spiking diffusion framework that integrates lateral connection and spiking neurons for image generation. The framework operates through two nested time-step cycles: the outer loop with a diffusion denoising process and the inner loop that generates spike sequences via the SNN. We denote the outer loop time step for diffusion denoising as \(t_d\) and the inner loop time step for spiking dynamics as \(t_s\). Additionally, to ensure the clarity of mathematical \textbf{notation}, we will use \textbf{superscripts} to represent the inner loop time steps of the SNN, while \textbf{subscripts} will be determined based on the context, typically representing the diffusion time steps (e.g., \( t_d \)) or the layers of the network (e.g., \( r \)).

The loss function used in the denoising process is adopted from DDPM~\cite{ho2020denoising}:
\begin{equation}
    \mathcal{L}_{noise, t_d} := \mathbb{E}_{\boldsymbol{x}_0, \epsilon \sim \mathcal{N}(0,1)} \left\| \epsilon - \epsilon_{\theta}(\boldsymbol{x}_{t_d}, t_d) \right\|_2^2, \quad t_d=1:T_d
\end{equation}

where \( \boldsymbol{x}_{t_d} \) is the noisy input at step \(t_d\), generated by adding noise to the initial stimulus \( \boldsymbol{x}_0 \) as:

\begin{equation}
    \boldsymbol{x}_{t_d} = \sqrt{\alpha_{t_d}} \boldsymbol{x}_0 + \sqrt{1-\alpha_{t_d}} \epsilon 
\end{equation}

Here, \(\alpha_{1:T_d}\) is a predefined decreasing sequence in the range $(0,1]$, representing the noise variance, and \( \epsilon \sim \mathcal{N}(0,\boldsymbol{I}) \) is standard Gaussian noise. 

The proposed model incorporates a substructure selection module that dynamically selects subnetworks based on diffusion and spiking time steps. Assuming the network has \( L \) layers with substructure selection modules, \( \{\hat{\boldsymbol{l}}^{t_s}_k\}^{L}_{k=1} \) represents the output logits of the \( L \) corresponding substructure selection modules, which uniquely determine the forward structure of the network. For noisy input \( x_{t_d} \), the SNN encoder $\mathcal{E}$ first encodes it into \( T_s \) spikes $\{\tilde{I}^{t_s}\}$:
\begin{equation}
    \tilde{I}^{1},...\tilde{I}^{T_s} = \mathcal{E}(\boldsymbol{x}_{t_d}) \label{eq:spike_encode}
\end{equation}
Then, the model estimates the added noise \( \epsilon_{\theta} \) using the SNN decoder:
\begin{align}
    & O^{t_s} = f^{t_s}\left( \tilde{I}^{t_s}, \{\hat{\boldsymbol{l}}^{t_s}_k\}^{L}_{k=1} \right) \label{eq:Os} \\
    & \epsilon_{\theta} = \mathcal{D}_s \left( \{ O_{t_s} \}^{T_s}_{t_s=1} \right) \label{eq:denoise}
\end{align}

In Equation~\ref{eq:Os}, \( O^{t_s} \) represents the network's output at the spike time step \( t_s \), \( L \) denotes the total number of network layers with the substructure selection module, and \( f^{t_s} \) refers to the proposed spiking model at \( t_s \). \( f^{t_s} \) takes two parameters: the input spike \( \tilde{I}^{t_s} \), and the logits of the substructure selection network across the \( L \) network layers, \( \{\hat{\boldsymbol{l}}^{t_s}_k\}^{L}_{k=1} \), which indicate the selected substructures at the time step \( t_s \). In Equation~\ref{eq:denoise}, the spike-to-static decoder \( \mathcal{D}_s \) averages the outputs at each spiking time step \( t_s \) to obtain the final predicted noise.


We use lateral connection between neurons to iteratively refine the output of the substructure selection network and \( \{\hat{\boldsymbol{l}}^{t_s}_k\}^{L}_{k=1} \). The update rules are as follows:

\begin{align}
        & \boldsymbol{l}^{t_s}_r = S_r \left( \boldsymbol{e_{t_d}} \right) \label{eq:lateral_1} \\
	& \boldsymbol{m}^{t_s+1}_r, \boldsymbol{\delta}^{t_s}_r = \gamma \cdot \Omega_r \left( \boldsymbol{\Lambda}_{\text{agg}} \boldsymbol{l}^{t_s}_r, \boldsymbol{m}^{t_s}_r \right) \label{eq:lateral_2}\\
	& \boldsymbol{\hat{l}}^{t_{s}}_r = \boldsymbol{l}^{t_s}_r - \boldsymbol{\delta}^{t_s}_r \label{eq:lateral_3}
\end{align}

\begin{figure}[t]
    \centering
    \includegraphics[width=1.0\columnwidth]{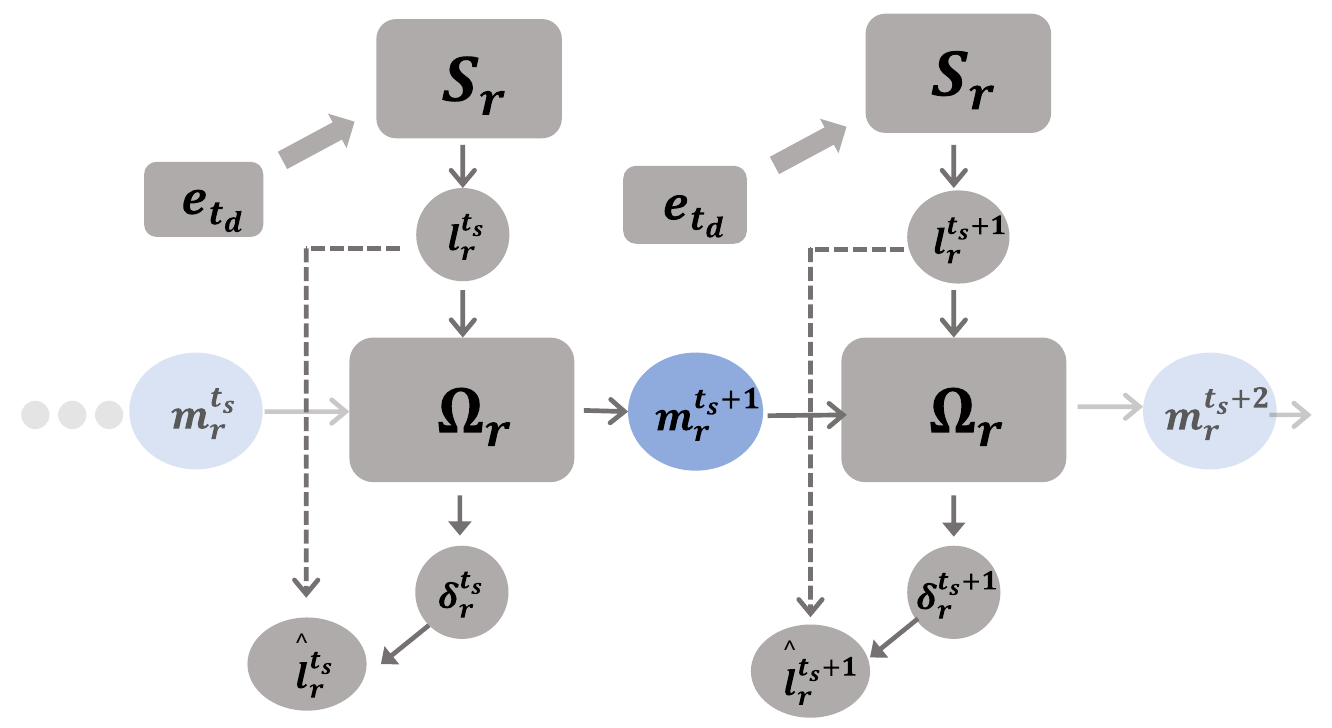}
    \caption{The schematic diagram of the interactions between the states of the variables in the lateral connection update process in Equation~\ref{eq:lateral_1},~\ref{eq:lateral_2} and~\ref{eq:lateral_3}. This update process is similar to a model with recurrent units, where the membrane potential \( m^{t_s}_r \) can be viewed as a recurrent hidden state.}
    \label{fig:state_diagram}
\end{figure}

In the above equations, \( \boldsymbol{l}^{t_s}_r \) represents the uncorrected logits at layer \( r \) and time step \( t_s \), while \( \boldsymbol{\hat{l}}^{t_s}_{k} \) represents the logits after lateral correction. In Equation~\ref{eq:lateral_1}, \( S_r \) represents the substructure selection network at layer \( r \), which is responsible for outputting the initial \( \boldsymbol{l}^{t_s}_{k} \). \( \boldsymbol{e_{t_d}} \) is a learnable embedding set at the diffusion time step \( t_d \). In Equation~\ref{eq:lateral_2}, \( \Omega_{r} \) represents the lateral mapping function at layer \( r \), which is governed by the dynamics of LIF neurons. \( \Omega_{r} \) takes two parameters: \( \boldsymbol{m}^{t_s}_r \), the membrane potential of the LIF neurons in $\Omega_r$ at time step \( t_s \), and \( \boldsymbol{\Lambda}_{\text{agg}} \boldsymbol{l}^{t_s}_r \), the logits adjusted by the aggregation matrix \( \boldsymbol{\Lambda}_{\text{agg}} \). The use of \( \boldsymbol{\Lambda}_{\text{agg}} \) is to model the interaction between each element of \( \boldsymbol{l}^{t_s}_r \), i.e., the interaction between output heads. \( \Omega_r \) outputs the lateral correction value \( \delta^{t_s}_r \) and the membrane potential \( \boldsymbol{m}^{t_s+1}_r \) for the next spike time step. \( \gamma \) is a learnable factor that controls the strength of the lateral connection update. In Equation~\ref{eq:lateral_3}, we use the lateral correction value \( \delta^{t_s}_r \) along with \( \boldsymbol{l}^{t_s}_r \) to obtain the corrected logits \( \boldsymbol{\hat{l}}^{t_s}_{k} \).

After the correction process described above, we convert the logits into probabilities \( \mathcal{P}_r = \text{Softmax}(\hat{\boldsymbol{l}}^{t_s}_{r,1}, \dots, \hat{\boldsymbol{l}}^{t_s}_{r,N}) \) for the \( N \) candidate subnetworks. The top \( k \) subnetworks are selected using a thresholding mechanism:

\begin{align}
& \boldsymbol{h} = \text{TopK} \left( \mathcal{P}_r, k \right) \\
& \text{Output} ( \boldsymbol{z} ) = \sum_{j=1}^{N} \boldsymbol{h}_j \cdot \mathcal{S}_j(z)
\end{align}

where \( \boldsymbol{h} \) is a masked probability vector, and \( \{\mathcal{S}_1, \dots, \mathcal{S}_N\} \) are the \( N \) candidate subnetworks. 

For the update process in Equation~\ref{eq:lateral_1},~\ref{eq:lateral_2} and~\ref{eq:lateral_3}, it can be viewed as a model with recurrent units, where the membrane potential \( m^{t_s} \) serves as the recurrent latent variable. Since the updates occur between different output neurons, it is independent of the standard forward propagation process and is thus considered an update based on lateral connections. 
In Figure~\ref{fig:pipeline}, we present an overview of the model implementation, highlighting the general process. To provide a clearer description, particularly regarding the state changes during lateral updates, we refer to Figure~\ref{fig:state_diagram}, which illustrates the relationships between the various factors in the lateral connection update process.

\begin{table}[t]
  \caption{The initialization with different settings for lateral factors and aggregation matrix, and the resulting FID.} \label{tab:param_init}
\centering
\resizebox{0.6\columnwidth}{!}{ 
\begin{tabular}{ccc}
\toprule
$\gamma^{\text{init}}$ & \(\boldsymbol{\Lambda}^{\text{init}}_{\text{agg}}\)                                     & FID$\downarrow$ \\ \midrule
\multirow{4}{*}{0.0}   & Additive Inverse                                                         & 2.31            \\
                       & Scaled                                                                   & 2.40            \\
                       & Inverse Scaled                                                           & 2.54            \\
                       & Random Binary                                                            & 2.45            \\ \midrule
1.0                    & \multirow{5}{*}{\( \mathbf{1}_{N \times N} - \mathbb{I}_{N \times N} \)} & 2.98            \\
10.0                   &                                                                          & 2.86            \\
100.0                  &                                                                          & 3.39            \\
-1.0                   &                                                                          & 2.79            \\
-10.0                  &                                                                          & 2.82            \\ \midrule
\cellcolor{gray!20}0.0                    & \cellcolor{gray!20}\( \mathbf{1}_{N \times N} - \mathbb{I}_{N \times N} \)                  & \cellcolor{gray!20}\textbf{2.10}            \\ \bottomrule
\end{tabular}
}
\end{table}

Regarding the initialization of the lateral parameters, we set the learnable scaling factor $\gamma$ to 0, which prevents lateral propagation during the initial training phase, thereby ensuring the stability of the early training process. We initialize the information aggregation matrix \( \boldsymbol{\Lambda}_{\text{agg}} \) in the form of \( \mathbf{1}_{N \times N} - \mathbb{I}_{N \times N} \), where \( \mathbb{I}_{N \times N} \) denotes the \( N \times N \) identity matrix, and \( \mathbf{1}_{N \times N} \) denotes the \( N \times N \) matrix of all ones. This ensures that, during the early training phase, \(\boldsymbol{\Lambda}_{\text{agg}}\) maintains the interaction between neurons in a mutually inhibitory state. In Section~\ref{sec:analysis_lateral} of the experimental part, we conducted a comparative analysis of different initialization settings for $\gamma$ and \(\boldsymbol{\Lambda}_{\text{agg}}\), resulting in Table~\ref{tab:param_init}. The results show that the optimal FID is achieved with this initialization setting.

\subsection{Theoretical Similarities with Biologically Plausible Update Mechanisms} \label{sec:theory}

We establish the biological plausibility of the lateral connection mechanism in Section~\ref{sec:model} by formulating a mathematical model for neurons with lateral interactions. Specifically, we demonstrate that, under a local objective function, the learning process of a neuron population with lateral connections exhibits similarities to the STDP mechanism. 

To formalize this, we define a latent variable \( A_i(t) \) to represent the activity level of neuron \( i \) at time \( t \), subject to the following constraint:

\begin{align}
	&A_i(t) = \underbrace{b_i(t)}_{(a)} + \underbrace{\gamma \cdot \sum_{j} \Psi_{ji} \left( \rho_j \right)}_{(b)} \label{eq:constraint} \\
        &\rho_j = \rho \left( A_j(t) \right)
\end{align}

Here, \( b_i(t) \) denotes the baseline activity of neuron \( i \), while \( \gamma \) is a scaling factor. The term \( \rho_j = \rho(A_j(t)) \) represents the explicit activity pattern of neuron \( j \), mapped from its latent activity state by a function \( \rho \), which can be interpreted as the firing rate. The function \( \Psi_{ji}(\cdot) \) quantifies the influence of neuron \( j \) on neuron \( i \). In the implementation of Section~\ref{sec:model}, \(\Omega\) and \(\boldsymbol{\Lambda}_{\text{agg}}\) in Equation~\ref{eq:lateral_1} correspond to the interaction function \( \Psi \) in Equation~\ref{eq:constraint}.

This formulation can be decomposed into two terms: (a) the intrinsic activity level of neuron \( i \), independent of network interactions, and (b) the influence from other neurons via their firing patterns \( \rho(A_j(t)) \). Given a local loss function \( \mathcal{L}_{local} \), we optimize the lateral connection parameter \( \boldsymbol{\omega} \) to minimize \( \mathcal{L}_{local} \):

\begin{align}
	\frac{\partial \mathcal{L}_{local}}{\partial \boldsymbol{\omega} } &= \sum_i \frac{\partial \mathcal{L}_{local} }{\partial A_i} \frac{\partial A_i}{\partial \boldsymbol{\omega}} \\
	&= \sum_i \frac{\partial \mathcal{L}_{local}}{\partial A_i} \left( \underbrace{\frac{\partial b_i(t)}{\partial \boldsymbol{\omega}}}_{0} + \gamma \cdot \sum_j \frac{\partial \Psi_{ji}}{\partial \boldsymbol{\omega}} \right) \label{eq:stdp_2} \\
	&= \gamma \cdot \sum_i \sum_j \frac{\partial \mathcal{L}_{local}}{\partial A_i} \frac{\partial \Psi_{ji}}{\partial \boldsymbol{\omega}} 
\end{align}

Since \( b_i(t) \) represents the baseline activity of the neuron, originating from forward propagation, we assume that it is independent of the lateral parameter \( \boldsymbol{\omega} \). We set \( \frac{\partial b_i(t)}{\partial \boldsymbol{\omega}} = 0 \). Using a first-order Taylor expansion, we approximate:

\begin{align}
	\frac{\partial \Psi_{ji}}{\partial \boldsymbol{\omega}} &\approx  \frac{\partial \left( \Psi_{ji}(0)+\Psi^{(1)}_{ji}(0) \cdot \rho_j \right) }{\partial \boldsymbol{\omega}} \label{eq:approxf_1} \\
	&= \frac{\partial \Psi^{(1)}_{ji}(0) \cdot \rho_j }{\partial \boldsymbol{\omega}}
\end{align}

Substituting into Equation~\ref{eq:stdp_2}, we obtain:

\begin{equation}
	\frac{\partial \mathcal{L}_{local}}{\partial \boldsymbol{\omega}} \approx \gamma \cdot \sum_i \frac{\partial \mathcal{L}_{local}}{\partial A_i} \frac{\partial \Psi^{(1)}_{ji}(0)}{\partial \boldsymbol{\omega}} \cdot \rho_j \label{eq:approx_fr}
\end{equation}

For an appropriate choice of \( \mathcal{L}_{local} \), the resulting update rule aligns with STDP. For instance, if we minimize the entropy of neuronal activity distributions:

\begin{equation}
	\mathcal{L}_{local} = - \sum_k p_k \log p_k 
\end{equation}

where \( p_k \) is a function of \( A_i \). We assume that \( p_k \) is positively correlated with \( A_k \), then:

\begin{equation}
	\frac{\partial \mathcal{L}_{local}}{\partial A_i} \simeq  -A_i - \sum_{j \neq i} (A_j \cdot \frac{\partial A_j}{\partial A_i}) \label{eq:prob}
\end{equation}

Since \( \frac{\partial A_j}{\partial A_i} \) is often negative~\cite{kullmann2012plasticity,froemke2015plasticity}, this suggests a competitive interaction where increased activity in one neuron inhibits others. Substituting into Equation~\ref{eq:approx_fr}, we derive:

\begin{equation}
	\frac{\partial \mathcal{L}_{local}}{\partial \boldsymbol{\omega}} \approx \gamma \cdot \sum_i \underbrace{ \left( A_i + \sum_k A_k \frac{\partial A_k}{\partial A_j} \right) }_{(a)} \cdot  \underbrace{-\frac{\partial \Psi^{(1)}_{ji}(0)}{\partial \boldsymbol{\omega}}}_{(b)} \cdot \underbrace{\rho_j}_{(c)} \label{eq:main}
\end{equation}

Each term in Equation~\ref{eq:main} has an intuitive interpretation: (a) Positively correlates with neuron \( i \)’s activity. (b) Represents the negative gradient of the interaction function with respect to \( \boldsymbol{\omega} \), assumed positive without loss of generality. (c) Corresponds to neuron \( j \)’s firing rate, which reflects its activity.

Notably, this formulation indicates that when neuron \( j \) fires synchronously with neuron \( i \), the parameter update is more pronounced. This mirrors the ``fire together, wire together" principle, highlighting a connection between lateral connection and STDP-like synaptic modifications. Thus, our model naturally aligns with biologically plausible learning dynamics.

\section{Experiments}
\subsection{Setup}
For our experimental setup, we partition the input data into patches of size 2, which are subsequently processed by a spiking transformer block. The transformer architecture consists of 12 layers, a hidden dimension of 768, and 12 attention heads. For smaller datasets, such as those in the MNIST series, we reduce both the hidden dimension and the number of attention heads by half.

Each model is trained for up to 800k steps, and the best-performing checkpoint is selected for evaluation. The learning rate is fixed at \(1 \times 10^{-4}\) without employing any learning rate scheduling. For substructure selection, each transformer block is assigned three candidate substructures, with two selected as the output during training.

In the outer loop of the diffusion process, we adopt 1000 time steps, whereas the inner loop of the SNN operates with two time steps per iteration. For evaluation, we employ a 1000-step denoising process to generate images and assess performance using either 5k or 10k generated samples as the benchmark.
\begin{figure}[t]
    \centering
    \includegraphics[width=1.0\columnwidth]{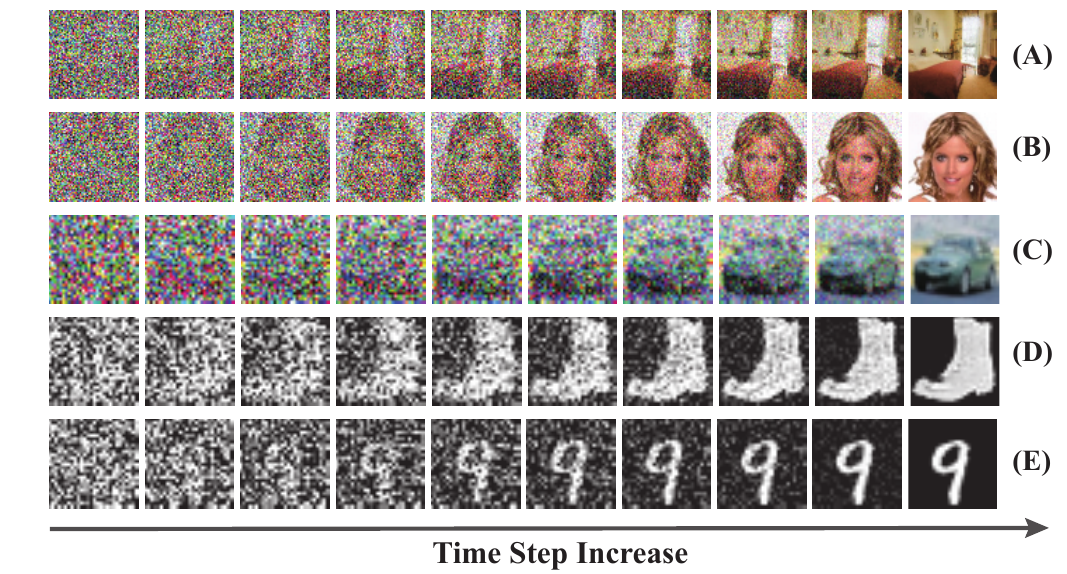}
    \caption{Denoising process of the proposed model on (A) LSUN Bedroom, (B) CelebA, (C) CIFAR-10, (D) Fashion MNIST, and (E) MNIST datasets.}
    \label{fig:denoising}
\end{figure}
\subsection{Comparative Analysis with SNN-based Generative Models}
We evaluate our method against state-of-the-art SNN-based generative models on CIFAR-10, CelebA, $64 \times 64$ LSUN Bedroom~\cite{yu2015lsun}, MNIST, and FashionMNIST datasets. The benchmarked models include FSVAE~\cite{kamata2022fully}, SGAD~\cite{feng2024spiking}, SDDPM~\cite{cao2024spiking}, SDiT~\cite{yang2024sdit}, and Spiking-Diffusion~\cite{liu2023spiking}.

\begin{figure*}[h]
    \centering
    \includegraphics[width=1.0\textwidth]{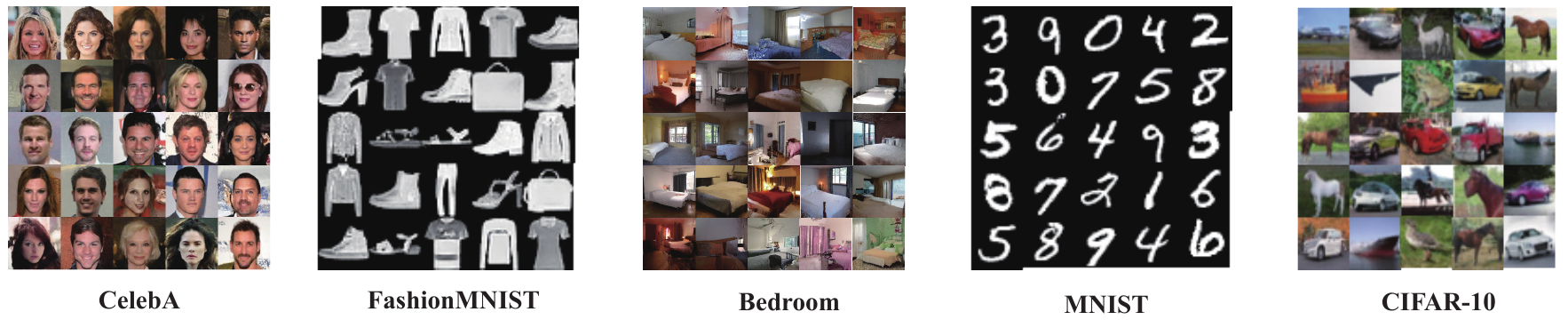}
    \caption{Visualization of generated images on CelebA, FashionMNIST, LSUN Bedroom, MNIST and CIFAR-10 datasets}
    \label{fig:show_img}
\end{figure*}

To quantify the generative performance, we compute the Fréchet Inception Distance (FID)\cite{yu2021frechet} for each approach. As presented in Table\ref{tab:res}, our method consistently achieves superior FID scores across most datasets, demonstrating its effectiveness in generating high-quality samples. 
To visually demonstrate the performance of the proposed method, we present sampled images from various datasets in Figure~\ref{fig:show_img}. Additionally, in Figure~\ref{fig:denoising}, we illustrate the reverse denoising process as the diffusion time steps increase.

\begin{table*}[t]
	\caption{Comparison of FID scores across various datasets with other SNN-based generative models.}
	\label{tab:res}
	\centering
\begin{tabular}{@{}ccccc@{}}
\toprule
\textbf{Dataset}                    & \textbf{Resolution}             & \textbf{Model}                          & \textbf{Time Steps} & \textbf{FID} $\downarrow$ \\ \midrule
\multirow{4}{*}{CelebA$^{*}$}       & \multirow{4}{*}{$64 \times 64$} & FSVAE~\cite{kamata2022fully}            & 16                  & 101.60                    \\
                                    &                                 & SGAD~\cite{feng2024spiking}                & 16                  & 151.36                    \\
                                    &                                 & SDDPM~\cite{cao2024spiking}             & 4                   & 25.09                     \\
                                    &                                 & \textbf{Ours}                           & 2                   & \textbf{19.35}            \\ \midrule
\multirow{2}{*}{LSUN Bedroom$^{*}$} & \multirow{2}{*}{$64 \times 64$} & SDDPM~\cite{cao2024spiking}             & 4                   & 47.64                     \\
                                    &                                 & \textbf{Ours}                           & 2                   & \textbf{29.77}            \\ \midrule
\multirow{6}{*}{MNSIT$^{*}$}        & \multirow{6}{*}{$28 \times 28$} & FSVAE~\cite{kamata2022fully}            & 16                  & 97.06                     \\
                                    &                                 & SGAD~\cite{feng2024spiking}                & 16                  & 69.64                     \\
                                    &                                 & Spiking-Diffusion~\cite{liu2023spiking} & 16                  & 37.50                     \\
                                    &                                 & SDDPM~\cite{cao2024spiking}             & 4                   & 29.48                     \\
                                    &                                 & SDiT~\cite{yang2024sdit}                & 4                   & 5.54                      \\
                                    &                                 & \textbf{Ours}                           & 2                   & \textbf{2.10}             \\ \midrule
\multirow{6}{*}{FashionMNIST$^{*}$} & \multirow{6}{*}{$28 \times 28$} & FSVAE~\cite{kamata2022fully}            & 16                  & 90.12                     \\
                                    &                                 & SGAD~\cite{feng2024spiking}                & 16                  & 165.42                    \\
                                    &                                 & Spiking-Diffusion~\cite{liu2023spiking} & 16                  & 91.98                     \\
                                    &                                 & SDDPM~\cite{cao2024spiking}             & 4                   & 21.38                     \\
                                    &                                 & SDiT~\cite{yang2024sdit}                & 4                   & \textbf{5.49}             \\
                                    &                                 & \textbf{Ours}                           & 2                   & 5.90                     \\ \midrule
\multirow{6}{*}{CIFAR-10}           & \multirow{6}{*}{$32 \times 32$} & FSVAE~\cite{kamata2022fully}            & 16                  & 175.50                    \\
                                    &                                 & SGAD~\cite{feng2024spiking}                & 16                  & 181.50                    \\
                                    &                                 & Spiking-Diffusion~\cite{liu2023spiking} & 16                  & 120.50                    \\
                                    &                                 & SDDPM~\cite{cao2024spiking}             & 4                   & 16.98                     \\
                                    &                                 & SDiT~\cite{yang2024sdit}                & 4                   & 22.17                     \\
                                    &                                 & \textbf{Ours}                           & 4                   & \textbf{12.81}            \\ \bottomrule
\end{tabular}
\end{table*}

\subsection{Ablation Study}

To evaluate the effectiveness of the proposed substructure selection mechanism and lateral connection method, we conducted an ablation study by performing incremental comparisons. Specifically, we analyzed the Fréchet Inception Distance (FID) scores of models with different configurations relative to the baseline, as summarized in Table~\ref{tab:ablation}. Here, \textbf{Baseline} denotes the diffusion model utilizing a spiking transformer, \textbf{Selection} refers to the addition of the substructure selection module, and \textbf{Lateral} represents the inclusion of lateral connections on top of \textbf{Selection}. The results clearly demonstrate that incorporating the substructure selection mechanism and lateral connection significantly improves model performance, confirming the effectiveness of our approach.

\begin{table}[t]
	\caption{Comparison of FID scores for models employing different methods on CIFAR-10, CelebA, and LSUN Bedroom datasets. The results demonstrate that the proposed methods substantially enhance model performance.}
	\label{tab:ablation}
	\centering
\begin{tabular}{@{}ccccc@{}}
\toprule
\multirow{2}{*}{\textbf{Dataset}} & \multicolumn{3}{c}{\textbf{Methods}}                      & \multirow{2}{*}{\textbf{FID $\downarrow$}} \\ \cmidrule(lr){2-4}
                                  & \textbf{Baseline} & \textbf{Selection} & \textbf{Lateral} &                                            \\ \midrule
\multirow{3}{*}{CIFAR-10}         & \checkmark        &                    &                  & 21.28                                      \\
                                  & \checkmark        & \checkmark         &                  & 13.85                                      \\
                                  & \checkmark        & \checkmark         & \checkmark       & \textbf{12.81}                             \\ \midrule
\multirow{3}{*}{CelebA}           & \checkmark        &                    &                  & 34.83                                      \\
                                  & \checkmark        & \checkmark         &                  & 31.31                                      \\
                                  & \checkmark        & \checkmark         & \checkmark       & \textbf{19.35}                             \\ \midrule
\multirow{3}{*}{Bedroom}          & \checkmark        &                    &                  & 178.14                                     \\
                                  & \checkmark        & \checkmark         &                  & 332.83                                     \\
                                  & \checkmark        & \checkmark         & \checkmark       & \textbf{29.77}                             \\ \bottomrule
\end{tabular}
\end{table}

\begin{table}[!t]
\caption{FID scores for different configurations of the total number of substructures \(N\) and the final number of selected substructures \(k\).}
\label{tab:num_experts}
\centering
\resizebox{1.0\columnwidth}{!}{ 
\begin{tabular}{@{}ccccc@{}}
\toprule
\textbf{FID} $\downarrow$ & $N=2,k=1$ & $N=3,k=2$ & $N=5,k=2$ & $N=7,k=2$ \\ \midrule
MNIST           & 2.657     &  \textbf{2.102}     & 2.587     & 2.663     \\
FashionMNIST    & 6.229     &  \textbf{5.902}     & 7.389     & 6.599     \\ \bottomrule
\end{tabular}
}
\end{table}

To further examine the influence of the total number of candidate substructures $N$ in the selection module and the final selected number $k$, we conducted additional experiments on the MNIST and FashionMNIST datasets. The corresponding FID scores under different configurations are presented in Table~\ref{tab:num_experts}. The results indicate that while increasing $N$ provides greater flexibility in substructure selection, it does not necessarily translate to improved generative performance. Instead, we observe that the configuration  \(N=3, k=2\) consistently yields the best trade-off between performance and model complexity. Given these findings, we employ this setting in subsequent experiments to ensure an optimal balance between computational efficiency and generative quality.



\begin{figure*}[!t]
    \centering
    \includegraphics[width=0.8\textwidth]{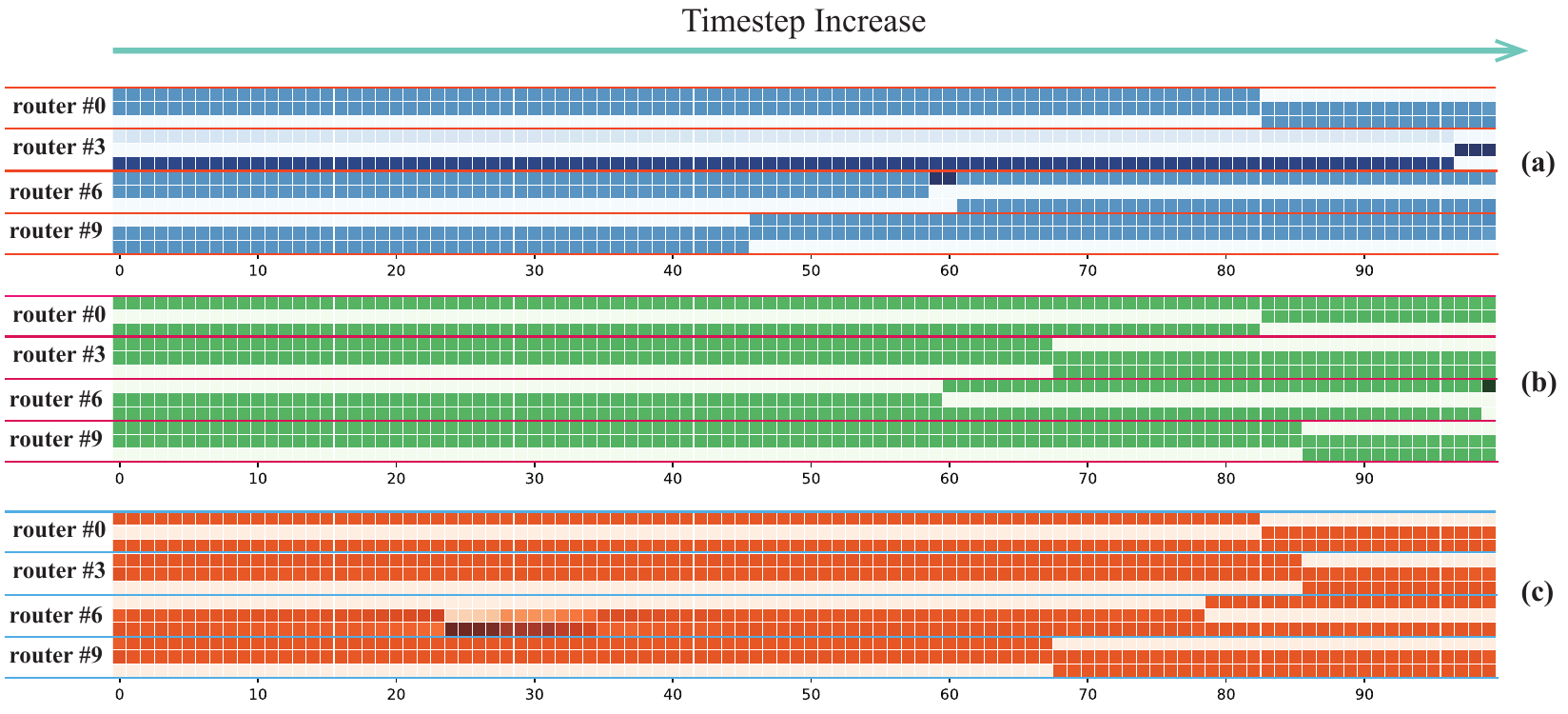}
    \caption{Visualization of substructure selection probabilities across different layers under the \(N=3, k=2\) configuration. Subfigures (a), (b), and (c) correspond to results on CIFAR-10, MNIST, and FashionMNIST, respectively. The selection varies with both layer depth and diffusion time step, with darker colors indicating higher selection probabilities.}
    \label{fig:which_expert}
\end{figure*}

To analyze the dynamic properties of the selection mechanism, we visualized the probability distributions of substructure selection across different network layers. As shown in Figure~\ref{fig:which_expert}, subfigures (a), (b), and (c) respectively present the results on CIFAR-10, MNIST, and FashionMNIST. The labels "router \#0, \#3, \#6, \#9" correspond to the selection probabilities of substructures at network layers 0, 3, 6, and 9, where darker colors indicate higher selection probabilities. From the visualizations, we observe that the selected substructures dynamically change with the input diffusion time step \( t_d \) (horizontal axis). This indicates that the model adapts its selection strategy at different diffusion stages, employing different substructure combinations to accommodate varying levels of denoising. This result further validates the dynamic adaptability of the selection mechanism, demonstrating that the model can autonomously adjust its computational pathways to enhance both generative quality and efficiency.

\subsection{Substructure Positioning in Network Layers}
To investigate the impact of expert positioning within the network, we implemented a multi-layer spiking-based Transformer where our proposed substructure selection mechanism was incorporated across all layers. In this experiment, we systematically examined how the placement of our module at different depth levels—shallow (head), middle, and deep (tail)—affects denoising performance. The shallow layers primarily capture fundamental features that significantly impact generative quality, while the middle layers facilitate feature transformation, and the deep layers refine fine details. We conducted experiments on FashionMNIST and MNIST datasets under different training schedules and analyzed the performance of different positioning strategies. The training settings (i)-(iv) and (a)-(d) correspond to different training steps ranging from 100k to 400k, with the results summarized in Table~\ref{tab:control_experts}.
\begin{table}[!t]
\caption{FID under different training settings when incorporating the substructure selection network at various layer positions in the model.}
\label{tab:control_experts}
\centering
\begin{tabular}{@{}cccccc}
\toprule
\multirow{2}{*}{\textbf{Dataset}}      & \multirow{2}{*}{\textbf{Settings}} & \multicolumn{4}{c}{\textbf{FID} $\downarrow$}                    \\ \cmidrule(l){3-6} 
                                       &                                    & \textbf{head} & \textbf{middle} & \textbf{tail} & \textbf{full} \\ \midrule
\multirow{4}{*}{FashionMNIST} & (i)                                & 7.02          & 8.20            & 7.81          & 7.08          \\
                                       & (ii)                               & 6.76          & 7.62            & 10.73         & 7.06          \\
                                       & \cellcolor{green!20}(iii)                              & \cellcolor{green!20}6.31          & \cellcolor{green!20}6.83            & \cellcolor{green!20}6.44          & \cellcolor{green!20}\textbf{5.90}          \\
                                       & (iv)                               & 7.51          & 6.33            & 6.88          & 8.48          \\ \midrule
\multirow{4}{*}{MNIST}        & (a)                                & 2.75          & 3.01            & 3.22          & 2.51          \\
                                       & \cellcolor{green!20}(b)                                & \cellcolor{green!20}2.18          & \cellcolor{green!20}2.40            & \cellcolor{green!20}7.63          & \cellcolor{green!20}\textbf{2.10}          \\
                                       & (c)                                & 2.99          & 3.39            & 6.82          & 5.51          \\
                                       & (d)                                & 2.45          & 2.92            & 4.51          & 2.82          \\ \bottomrule
\end{tabular}
\end{table}
As shown in Table~\ref{tab:control_experts}, applying the substructure selection mechanism in the shallow layers (head) consistently leads to better generative performance compared to middle and deep layers. This suggests that early network stages play a crucial role in extracting structural information, which significantly influences the overall generation process. Furthermore, as training progresses, the impact of expert positioning changes. In earlier training stages (e.g., settings (i), (ii)), applying the mechanism in the deep layers does not yield optimal performance, whereas integrating it in the shallow layers improves results. In later training stages (e.g., settings (iii), (b)), placing the module across all layers achieves the best FID scores. However, when training continues beyond a certain point (e.g., settings (iv), (c)), we observe a degradation in performance, possibly due to overfitting.

These results highlight a critical trade-off between performance and computational efficiency. When computational resources are sufficient, full-network substructure selection can be employed to achieve the best generative performance. However, our experiments demonstrate that restricting the selection mechanism to the shallow layers (head) yields comparable results while significantly reducing computational costs. This suggests that for resource-constrained scenarios, applying substructure selection only to the early layers provides a cost-effective alternative, maintaining strong performance while improving efficiency. This flexibility makes our approach adaptable to a wide range of practical applications, balancing accuracy and efficiency based on available computing resources.


\begin{figure*}[t]
    \centering
    \includegraphics[width=1.0\textwidth]{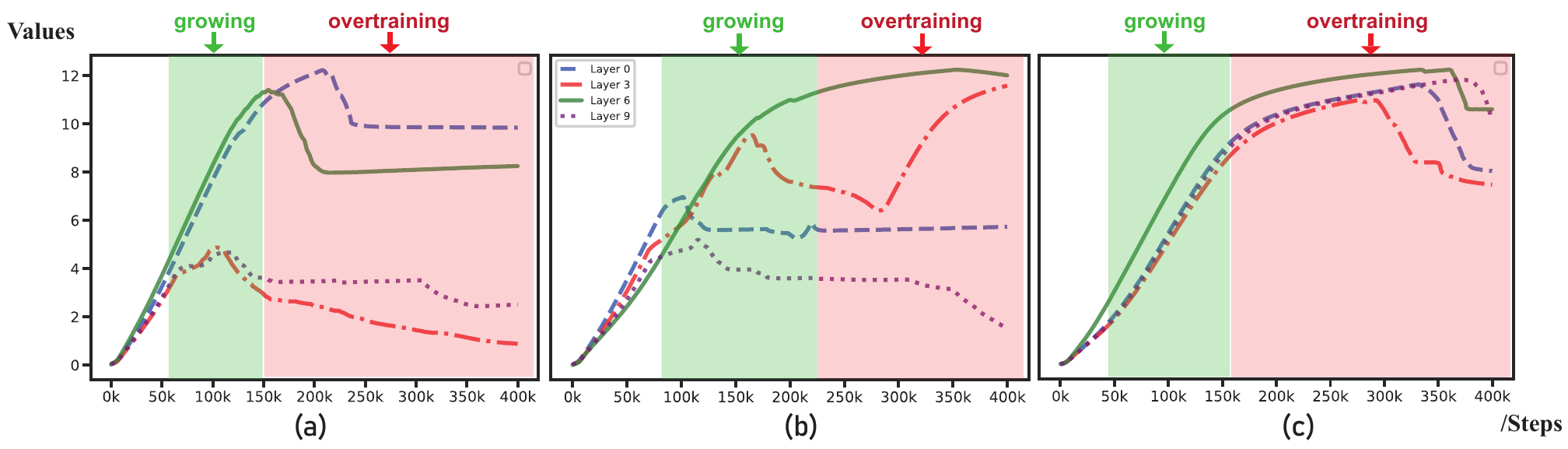}
    \caption{(a), (b), and (c) represent the variation of the lateral factor \(\gamma\) at different layers over the course of training steps on the MNIST, FashionMNIST, and CIFAR-10 datasets, respectively. During the growing phase, \(\gamma\) steadily increases, indicating an enhancement of lateral connections. In the overtraining phase, the model exhibits excessive training, leading to fluctuations in \(\gamma\).}
    \label{fig:lateral_factor}
\end{figure*}

\begin{figure*}[t]
    \centering
    \includegraphics[width=0.8\textwidth]{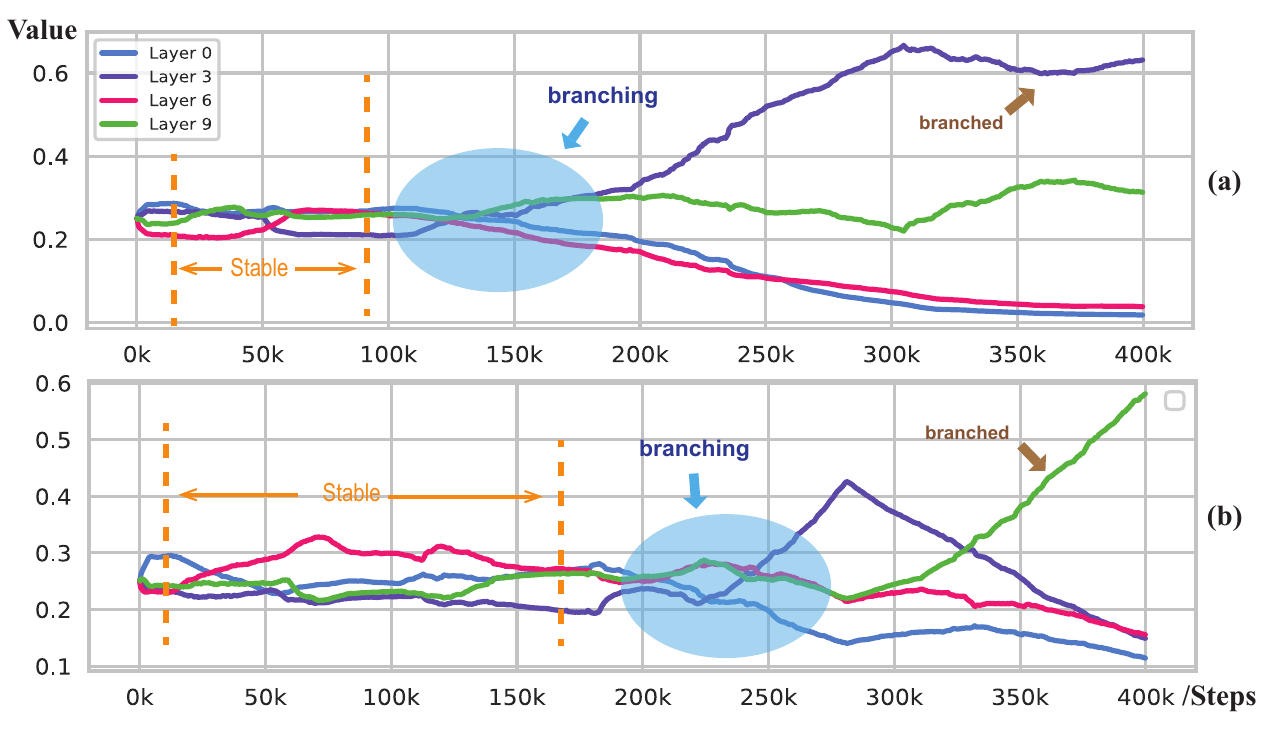}
    \caption{(a) and (b) represent the variation in the firing thresholds of the spiking neurons in \(\Omega\) on the MNIST and FashionMNIST datasets, respectively. The threshold exhibits differentiation across different layers.}
    \label{fig:threshold}
\end{figure*}

\begin{figure*}[!t]
    \centering
    \includegraphics[width=1.0\textwidth]{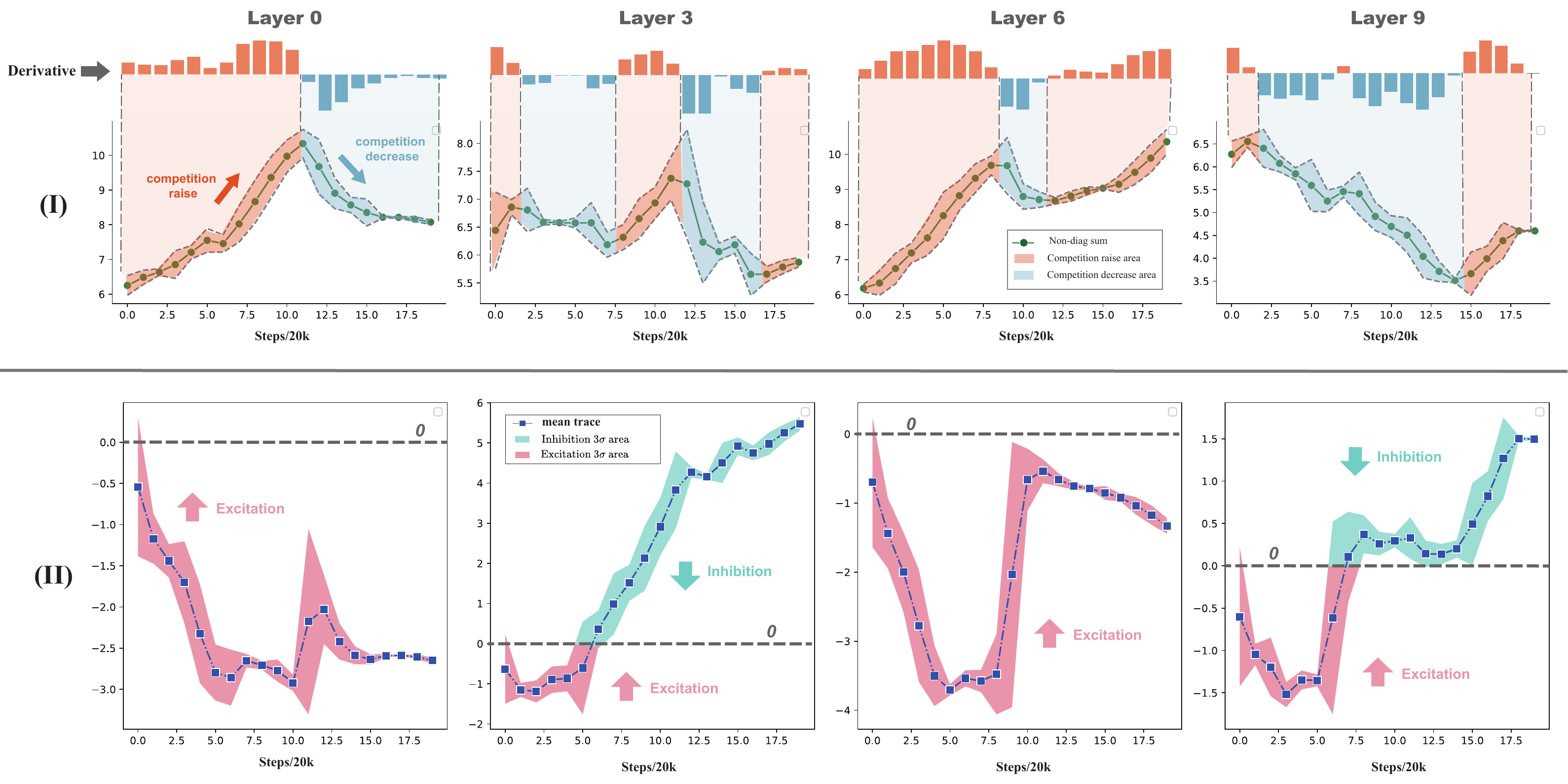}
    \caption{(I) Sum of the off-diagonal elements and their derivative of $\Lambda_{agg}$ at layers 0, 3, 6, and 9. (II) The trace of $\Lambda_{agg}$ at layers 0, 3, 6, and 9 }
    \label{fig:lateral_matrxi}
\end{figure*}
\subsection{Analysis of Lateral Parameters} \label{sec:analysis_lateral}

Equations~\ref{eq:lateral_1} and~\ref{eq:lateral_2} define key parameters governing lateral connection, including the lateral factor $\gamma$, the firing threshold of LIF neurons in the spike map \(\Omega\), and the aggregation matrix \(\Lambda_{agg}\). This section presents an analysis of their training dynamics and variations across different datasets.

Figure~\ref{fig:lateral_factor} illustrates the evolution of the lateral factor \(\gamma\) across network layers during training on the MNIST, FashionMNIST, and CIFAR-10 datasets, corresponding to subfigures (a), (b), and (c), respectively. The lateral factor exhibits a consistently increasing trend, indicating a progressive enhancement of lateral connections. This process can be categorized into a \textbf{growth phase} (green) and an \textbf{overtraining phase} (red). The optimal model performance occurs at the transition between these two phases, where lateral connectivity is at its peak.

Subfigures (a) and (b) of Figure~\ref{fig:threshold} illustrate the evolution of spiking neuron firing thresholds in \(\Omega\) across different layers during training. Initially, these thresholds remain stable but later differentiate based on layer depth. In \(\Omega\), the firing threshold can be interpreted as a preference for lateral connectivity. The results suggest that different layers exhibit distinct sensitivities to lateral interactions. Notably, model performance peaks when firing thresholds in $\Omega$ start to diverge (as marked in the figure).

The aggregation matrix $\Lambda_{agg}$ encodes interaction dynamics among neurons. We analyze it through its trace and the sum of its off-diagonal elements. Intuitively, the trace of $\Lambda_{agg}$ quantifies self-regulation in lateral neuron corrections: a \textbf{negative trace} indicates self-excitation, while a \textbf{positive trace }represents self-inhibition. The sum of the off-diagonal elements reflects the degree of lateral influence exerted by neighboring neurons, indicating either excitatory or inhibitory interactions.

Figure~\ref{fig:lateral_matrxi} (I) visualizes the sum of off-diagonal elements and their derivatives for $\Lambda_{agg}$ at layers 0, 3, 6, and 9 during training on the MNIST dataset. A \textbf{positive derivative} (red) signifies increasing mutual inhibition, while a \textbf{negative derivative} (blue) indicates reduced inhibition. Figure~\ref{fig:lateral_matrxi} (II) depicts the trace of $\Lambda_{agg}$, where red regions correspond to self-excitation and blue regions to self-inhibition. In layers 0 and 6, the off-diagonal sum exhibits an upward trend, while the trace suggests increasing self-excitation. These observations indicate that lateral connections contribute to the formation of dominant neurons, consistent with the theoretical insights from Equation~\ref{eq:main}.

Beyond tracking lateral parameter dynamics during training, we further investigated different initialization schemes. Specifically, we examined various initial values for \(\gamma\) (\{1, 10, 100, -1, -10\}) and compared them with the default setting \(\gamma = 0\). A large positive \(\gamma\) implies a strong reliance on lateral connections in early training, whereas a negative \(\gamma\) induces reversed updates. Additionally, we evaluated different initialization strategies for $\Lambda_{agg}$ relative to the default setting \(\mathbf{1}_{N \times N} - \mathbb{I}_{N \times N}\), including: \textbf{Additive Inverse}: Negating matrix values to eliminate initial mutual suppression; \textbf{Scaled}: Multiplying the matrix by a large factor (100); \textbf{Inverse Scaled}: Applying``Additive Inverse" followed by scaling; \textbf{Random Binary}: Initializing matrix elements randomly to 0 or 1.

Table~\ref{tab:param_init} presents the corresponding FID results, demonstrating that the optimal performance is obtained with \(\gamma = 0\) and $\Lambda_{agg} = \mathbf{1}_{N \times N} - \mathbb{I}_{N \times N}$.

\section{Conclusion}

This work presents a novel approach that integrates lateral connections with a substructure selection network to develop an SNN-based diffusion model. By formulating the interactions within lateral neural populations, we theoretically demonstrate that, under an appropriate local objective, the iterative refinement of the substructure selection network—facilitated by the spiking inner loop, lateral connection mechanism, and surrogate gradient—aligns with biologically plausible learning principles.  Empirical evaluations on multiple benchmark datasets, including MNIST, CIFAR-10, CelebA, and LSUN Bedroom, reveal that the proposed model consistently outperforms existing SNN-based generative models. Moreover, extensive ablation studies validate the contributions of substructure selection and lateral connections, providing an in-depth analysis of their functional impact throughout the training process. These findings underscore the significance of lateral connection and structured subnetwork selection in improving the performance of SNN-based diffusion models. We envision that this work will foster a deeper understanding of the role of lateral connectivity and hierarchical substructure selection in SNNs with spiking inner loops, and anticipate that these biologically inspired mechanisms will serve as a foundation for future advancements in neuromorphic computing and generative modeling.


\bibliographystyle{plain}
\bibliography{ref.bib} 

\end{document}